# Reference Model of Multi-Entity Bayesian Networks for Predictive Situation Awareness

*Cheol Young Park*[a]*** and *Kathryn Blackmond Laskey*[a]

[a]*The Sensor Fusion Lab & Center of Excellence in C4I, George Mason University, USA*

A R T I C L E   I N F O

A B S T R A C T

During the past quarter-century, situation awareness (SAW) has become a critical research theme, because of its importance. Since the concept of SAW was first introduced during World War I, various versions of SAW have been researched and introduced. Predictive Situation Awareness (PSAW) focuses on the ability to predict aspects of a temporally evolving situation over time. PSAW requires a formal representation and a reasoning method using such a representation. A Multi-Entity Bayesian Network (MEBN) is a knowledge representation formalism combining Bayesian Networks (BN) with First-Order Logic (FOL). MEBN can be used to represent uncertain situations (supported by BN) as well as complex situations (supported by FOL). Also, efficient reasoning algorithms for MEBN have been developed. MEBN can be a formal representation to support PSAW and has been used for several PSAW systems. Although several MEBN applications for PSAW exist, very little work can be found in the literature that attempts to generalize a MEBN model to support PSAW. In this research, we define a reference model for MEBN in PSAW, called a PSAW-MEBN reference model. The PSAW-MEBN reference model enables us to easily develop a MEBN model for PSAW by supporting the design of a MEBN model for PSAW. In this research, we introduce two example use cases using the PSAW-MEBN reference model to develop MEBN models to support PSAW: a Smart Manufacturing System and a Maritime Domain Awareness System.

## 1. Introduction

Predictive Situation Awareness (PSAW), the ability to estimate current situations and predict aspects of a temporally evolving situation, requires reasoning about multiple sensors and targets over time. Furthermore, the number of entities and the relationships among them may be uncertain. For this reason, PSAW needs an expressive formal language for representing and reasoning about uncertain, complex, and dynamic situations. Multi-Entity Bayesian Networks (MEBN) [Laskey, 2008] is such an expressive language, and has been applied to PSAW systems [Laskey et al., 2000][Wright et al., 2002][Costa et al., 2005][Suzic, 2005][Costa et al., 2012][Park et al., 2014][Golestan, 2016][Li et al., 2017][Park et al., 2017]. In a recent review of knowledge representation formalisms for situation awareness, Golestan et al. [2016] recommended MEBN as having the most comprehensive coverage of features needed to represent complex situations. Patnaikuni et al., [2017] reviewed various applications using MEBN.

Several MEBN modeling methodologies have been researched for efficient development of MEBN models. (1) Guidance for the modelers who develop MEBN models is available in the form of a reference architecture [Haberlin et al., 2013] and a structured process [Carvalho, 2011] for constructing semantically rich models for reasoning under uncertainty in complex environments. (2) A generic framework for plan recognition using MEBN has been introduced [Suzic, 2005]. The framework focuses on estimating plans of multi-agent organizations through time using an ontology that supports MEBN and contains four main template random variables (i.e., utility, plan, state, and observation) to represent a situation for agents in the multi-agent organizations. In the framework, an agent aims to optimize its utility (e.g., gain and loss). Also, the agents may have plans for various situations to fulfill the objectives of the agents. According to the plans, there are several states for the agents and the states may be observed.

Nevertheless, the process of constructing a MEBN model for a new application remains challenging. Different applications of MEBN to PSAW tend to have similar goals and common model elements. Therefore, even guided by the reference architecture [Haberlin et al., 2013] and the structured modeling process [Carvalho, 2011], designing a new MEBN model for PSAW from the ground up is inefficient. Incorporating the common model elements into a reference model, "an abstract framework for understanding significant relationships among the entities of some environment [MacKenzie et al., 2006]", in a specific domain (e.g., PSAW) promises to significantly reduce the development time and cost, and also result in a more well-formed model. The Suzic framework [Suzic, 2005] does not provide specific elements to match the elements of MEBN (e.g., MFrag, RV, LPD function, and entity (see Section 2.1)). In MEBN modeling for PSAW, it is necessary to know what kinds of MFrags, RVs, LPD functions, and entities are needed. Thus, pre-defined candidates for these model elements can make the model building process more efficient.

---
* Corresponding author.
E-mail address: cparkf@gmu.edu

This paper defines a *PSAW-MEBN reference model*[*]. The model specifies reference MFrags, RVs, and entities which support the design of a MEBN model for PSAW. Such a MEBN model for PSAW is called a *PSAW-MTheory*. The PSAW-MEBN reference model should include the elements of the situation such as targets, sensors, activities, locations, time, etc, as well as it being capable of generalization to express a variety of situations for PSAW (e.g., PSAW for naval operation and PSAW for critical infrastructure protection). Also, the model specifies reference MFrags, RVs, and entities which support the design of a PSAW-MTheory designed to reason about PSAW questions (e.g., "How many military vehicles are we going to encounter?", "How high will the level of danger for an enterprise be?", and "Where will the event occur?").

Section 2 provides background information about MEBN, Data Fusion Model, and Situation Awareness (SAW). In Section 3, we introduce what PSAW is, analyze what properties of PSAW are, discuss what OODA (*Observe – Orient – Decide – Act*) is, and present how to interpret PSAW in terms of OODA to investigate possible elements of PSAW from OODA. In Section 4, the PSAW-MEBN reference model is presented with the reference entities, the reference RVs, and the reference MFrags. In Section 5, two use cases using the PSAW-MEBN reference model are presented. Finally, conclusions are presented and future research directions are discussed.

## 2. Background

In this section, we introduce Multi-Entity Bayesian Networks (MEBN) first. We, then, introduce the JDL Data Fusion Model and Situation Awareness (SAW) to provide initial concepts for PSAW. PSAW, a specialization of SAW that emphasizes predicting an evolving situation, will be discussed in Section 3.

### 2.1. Multi-Entity Bayesian Network

MEBN extends Bayesian Networks by incorporating with First-Order Logic and can address reasoning challenges for complex and uncertain situations. Various hypotheses for situations can be represented by MEBN (e.g., entities, attributes of entities, methods of entities, and relationships of entities). In MEBN, such hypotheses can be associated with probability

---

theory and reasoned by Bayesian inference. A MEBN model, called an *MTheory*, can represent a hypothesis. An MTheory (e.g., Fig. 1) is a collection of MEBN fragments (*MFrags*). An MFrag (e.g., a blue rounded box in Fig. 1) represents conditional probability distributions of resident random variables (RVs) (e.g., yellow rounded boxes in Fig. 1). Resident random variables can take ordinary variables as arguments. These arguments can be instantiated for different domain entities. By taking the domain entities, instances of the resident random variables can be generated. The instances are random variables of a common Bayesian network. Such a Bayesian network is called a *situation-specific Bayesian Network* (SSBN). Thus, an SSBN (e.g., Fig. 2) is efficiently constructed from an MTheory (e.g., Fig. 1) using domain entities associated with a situation using an SSBN algorithm [Laskey, 2008]. We define elements of MTheory more precisely. The following definitions are taken from [Laskey, 2008].

**Definition 2.1 (MTheory)** An *MTheory M*, or MEBN Theory, is a collection of MFrags that satisfies conditions given in [Laskey, 2008] ensuring the existence of a unique joint distribution over its random variables.

An MTheory is a collection of MFrags that defines a consistent joint distribution over random variables describing a domain. The MFrags forming an MTheory should be mutually consistent. To ensure consistency, conditions must be satisfied such as no-cycle, bounded causal depth, unique home MFrags, and recursive specification condition [Laskey, 2008]. No-cycle means that the generated SSBN will contain no directed cycles. Bounded causal depth means that depth from a root node to a leaf node of an instance SSBN should be finite. Unique home MFrags means that each random variable has its distribution defined in a single MFrag, called its home MFrag. Recursive specification means that MEBN provides a means for defining the distribution for an RV depending on an ordered ordinary variable from previous instances of the RV.

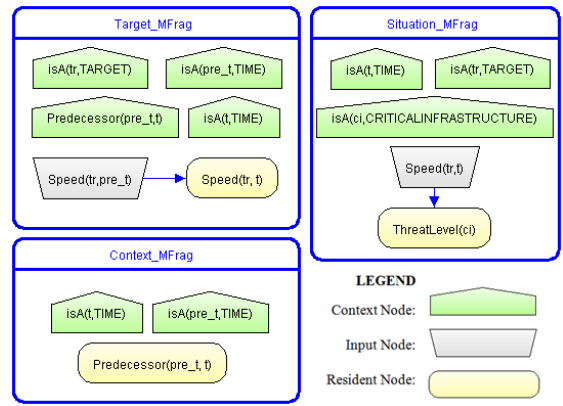

**Fig. 1 Simple Danger Assessment MTheory**

Fig. 1 shows an illustrative simple example of an MTheory used by a PSAW system for danger assessment. The MTheory can be used to predict the threat level (*High* and *Low*) of a critical infrastructure over time depending on the speed of a target approaching the critical infrastructure. It contains three MFrags: (1) *Target*, (2) *Situation*, and (3) *Context*. The MTheory also contains three kinds of entities: *Target*, *Critical Infrastructure*, and *Time*. The target entity represents a threatening entity which can attack a critical infrastructure. The critical infrastructure entity is an entity which the PSAW system defends. The time entity represents a time stamp during the situation flows.

**Definition 2.2 (MFrag)** An MFrag *F*, or MEBN fragment, consists of: (*i*) a set *C* of context nodes, which represent conditions under which the distribution defined in the MFrag is valid; (*ii*) a set *I* of input nodes, which have their distributions defined elsewhere and condition the distributions defined in the MFrag; (*iii*) a set *R* of resident nodes, whose distributions are defined in the MFrag[†]; (*iv*) an acyclic directed graph *G*, whose nodes are associated with resident and input nodes; and (*iv*) a set *L^C* of class local distributions, in which an element of *L^C* is associated with each resident node.

For example, in Fig. 1, the MFrag *Target* represents probabilistic knowledge of properties (i.e., speed) of targets. It contains three *isA* context nodes for entities *Target*, *Previous Time*, and *Time*. To determine a time relationship, the context node Predecessor($pre\_t$, $t$) is included in the MFrag. The node Predecessor($pre\_t$, $t$) means that the time interval $pre\_t$ occurs immediately before the time interval $t$. The MFrag, also, contains one resident node Speed($tr$, $t$) and one input node Speed($tr$, $pre\_t$) to represent the successive speeds of a target $tr$ over the timestamps $pre\_t$ and $t$. The nodes or RVs in an MFrag are different from the random variables in a common Bayesian network. We call the nodes in the MEBN nodes or MNodes.

**Definition 2.3 (MNode)** An *MNode*, or MEBN Node, is a random variable *N(ff)* specified an *n*-ary function or predicate of first-order logic, a list of *n* arguments consisting of ordinary variables, a set of mutually exclusive and collectively exhaustive possible values, and an associated class local distribution. The special values *true* and *false* are the possible values for predicates, but may not be possible values for functions. The RVs associated with the MNode are constructed by substituting domain entities for the *n* arguments of the function or predicate. The class local distribution specifies how to define local distributions for these RVs.

For example, the resident node Speed($tr$, $t$) in the MFrag *Target* is an MNode specified by a function of First-order logic as Speed($tr$, $t$). It has two possible values (i.e., *Fast* and *Slow*). This MNode is associated with the class local distribution, *L^C*. The MNode is used as a template for the distributions of instance random variables (e.g., $Speed\_tr1\_t1$ in Fig. 2) created when an SSBN is constructed from the MFrag associated with the MNode.

The MTheory in Fig. 1 can be used to construct various situation-specific Bayesian Networks (SSBNs) to support the PSAW system for danger assessment. As an example, consider a simple situation in which a target $tr1$ is approaching a critical infrastructure $ci1$ over three timestamps $t1$, $t2$, and $t3$ as shown Fig. 2.

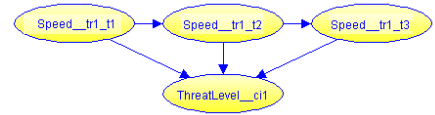

**Fig. 2 An SSBN from the Simple Danger Assessment MTheory**

This is just one example of the use of the MTheory (Fig. 1) to represent a danger assessment situation. We can use the MTheory to represent and reason about various danger assessment situations, each particular situation being represented by a different SSBN.

## 2.2. Data Fusion Model

In 1985, U.S. Joint Directors of Laboratories (JDL) Data Fusion Group proposed the Data Fusion Model (called JDL model) [White, 1988] which was revised to [Steinberg et al., 1998][Llinas et al., 2004][Liggins, et al., 2008]. JDL defines data fusion as "*Data fusion is a process dealing with the association, correlation, and combination of data and information from single and multiple sources to achieve refined position and identity estimates, and complete and timely assessments of situations and threats as well as their significance.* [White, 1991]". A concise definition for data fusion was presented as "*Data fusion is the process of combining data or information to estimate or predict entity states.* [Steinberg et al., 1998]".

Many definitions of data fusion, including the above two definitions, were reviewed and discussed in [Boström et al., 2007]. The process of data fusion means not only combining signals from low-level sensors but also combining knowledge from high-level sources to estimate and predict entity states. The process involves induction, deduction, and abduction, as does the thinking process of a human being. To implement the process, the JDL model provides four level stages as: "*Level 0: Signal/feature assessment. Estimation of signal or feature states. Signals and features may be defined as patterns that are inferred from observations or measurements. These may be static or dynamic and may have locatable or causal origins (e.g., an emitter, a weather front, etc.). Level 1: Entity assessment. Estimation of entity parametric and attributive states (i.e., of entities considered as individuals). Level 2: Situation assessment. Estimation of the structures of parts of reality (i.e., of sets of relationships among entities and their implications for the states of the related entities). Level 3: Impact assessment. Estimation of the utility/cost of signal, entity, or situation states, including predicted utility/cost given a system's alternative courses of action. Level 4: Process assessment. A system's self-estimation of its performance as compared to desired states and measures of effectiveness (MOEs).* [Liggins, et al., 2008]"

In Level 0, signal or feature states of an attribute of an entity are estimated from one or more signal observations (e.g., imagery, electromagnetic, and acoustic data) detected by sensors of various types. For example, from imagery data depicting movement of a vehicle, the size, velocity, shape or color of the moving vehicle can be estimated using a feature extraction method. The feature extraction method detects a pattern of signals or features (e.g., for the size of the moving vehicle, an image range from the moving vehicle in the imagery data can be regarded as a pattern), and it is mapped to a signal or feature state (e.g., small, medium, or large size of the moving vehicle).

---

[†] Bold italic letters are used to denote sets.

In Level 1, parametric and attributive states of entities (e.g., identity, location, track, and activity) considered as individuals are estimated from one or more signals, features, and entity states. The state can be considered as the state of a random variable or random vector. Example state variables include temperature, location, functional class, an attribute, or activity state of an entity.

In level 2, relationships or situations are estimated from one or more entity states and relationships. For estimation of a situation, an inference method is usually required. For example, Bayesian Networks [Steinberg, 2003] allow predicting a future situation based on current and historical situations.

In level 3, the impact of a signal, entity, or situation for a goal or mission is estimated. Impact assessment for threat assessment involves combining multiple sources of information to infer a conditional or counterfactual situation. Based on the estimated situation, known plans, and predicted reactions, outcome and cost in terms of one's goal or mission are estimated or predicted.

In level 4, measures of performance (MOPs) and measures of effectiveness (MOEs) of the system are estimated from the desired set of the system states and responses by performance analysis. Impact assessment of level 3 provides the impact to the goal or mission as technical performance measures (TPMs) in systems engineering. To assure of achievement of a goal, MOPs derived from MOEs and composed of TPMs should be estimated in level 4 [Liggins, et al., 2008].

### 2.3. Situation Awareness

The concept of SAW has been applied various domains which involve humans performing tasks in complex and dynamic systems. Examples include aviation [Jensen, 1997], air traffic control [Endsley & Smolensky, 1998], safety control [Salmon et al., 2006], automobile driving [Zheng et al., 2004], and C4I (Command, control, communication, computers, and intelligence) systems [Stanton et al., 2001]. In the literature related to SAW, Breton and Rousseau used a systematic approach to classify twenty-six SAW definitions [Breton & Rousseau, 2001]. They found a set of elements of SAW. They classified SAW into two top-level categories: State- and Process-oriented SAW. In their definition, Process-oriented SAW focuses on the link between the situation and the cognitive processes generating SAW, while State-oriented SAW focuses on the link between the situation and an internal representation of elements present in the situation. In the definitions of Process-oriented SAW, they include various processes such as perception, comprehension, projection, and action. In the definitions of State-oriented SAW, they classify awareness and knowing as a state of situation. According to the most commonly cited definition, SAW is composed of three processes;

"*Level 1: the perception of the elements in the environment within a volume of time and space, Level 2: the comprehension of their meaning, and Level 3: the projection of their status in the near future.* [Endsley, 1988; Endsley et al., 2003]"

In Level 1, the states, attributes, and dynamics of relevant elements in the environment should be detected accurately in a purpose of a system which uses SAW [Endsley et al., 2003]. A perception component of the system receives cues or signals from multiple sensors of the component. The sensors are separated from each other at this level. However, the sensors and their signals should be selected for the purpose of the system. In some cases, a critical sensor or signal is not present, so this level must ensure to perceive correct and necessary signals for the purpose of the system, and must compensate appropriately for degraded or missing inputs.

Level 2 is to understand what the signals mean in relation to the purpose of the system [Endsley et al., 2003]. For example, isolated words are perceived (Level 1). Then, by integrating the words in a certain manner, the individual words provide meaning to the sentence in which they occur (Level 2). By integrating signals, the signals form information. Then, the combined information is measured in how the information impacts the purpose of the system. As a result, a picture of the current situation showing relationships between information and impacts to the purpose is formed. At this level, understanding situation correctly is required.

Level 3 is to predict future states of the information in relation to the purpose of the system [Endsley et al., 2003]. For example, as a driver, by approaching a destination, the driver may estimate the distance between the current and destination location through time using her knowledge. Level 3 SAW is the estimation or prediction for the states of information through time. Time plays a significant role of understanding situation. A snapshot of situations is not enough. For example, we may want to know an answer about "how much time is available until some event occurs or some action must be taken?" [Endsley et al., 2003].

[Lambert, 2001] noted the similarity between levels 1 ~ 3 of the JDL model and Endsley's SAW model. Both models describe estimation and prediction of a situation as part of reality. Note, however, that Endsley's SAW model provides an abstraction of the processes of perception, comprehension, and projection, but does not provide specific processes of how to implement perception, comprehension, and projection, while the JDL model provides somewhat specific processes through levels 0 ~ 4.

## 3. Predictive Situation Awareness

### 3.1. Definition of PSAW

There are several definitions of a situation in the literature. Merriam-Webster Dictionary [Merriam-Webster's situation, 2015] defines it as "*the way in which something is placed in relation to its surroundings*". Wordnet Dictionary [Wordnet's situation, 2015] defines it as "*the general state of things; the combination of circumstances at a given time;*". Devlin [1995] defined a situation as a structured part of reality which is individuated by an agent. Sowa [2011] defined a situation as "*a region of space-time that bounds the range of immediate perception, action, interaction, and communication of one or more agents*". Devlin [2006] discussed the definition from Barwise & Perry [1981] in which a situation consists of "*objects having properties and standing in relations to one another* [Barwise & Perry, 1981]". A situation can be described as relations of objects existing in a certain circumstance at a given time, reflecting a partial world, and can be perceived by an agent.

Fig. 3 shows an illustrative example for Situation Awareness (SAW). A person observes the target situation in the world, in which two vehicles are moving. In the target situation, the first vehicle is actually a wheeled vehicle, while the second vehicle is actually a tracked vehicle. The person makes a picture of the interpreted situation in the person's mind, in which two wheeled vehicles are moving. The interpreted situation in the person's awareness and the target situation in the world are different, because of various internal and external factors (e.g., the person's knowledge about the vehicle, the person's eyesight, and the observing environment between the person and the vehicles). An overall situation containing the person, the environment of the person, the interpreted situation, and the target situation can be called a *perceived situation*.

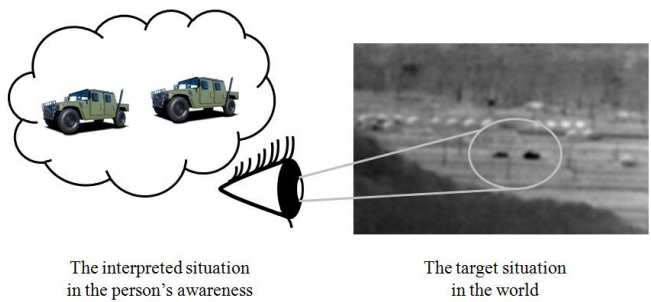

The interpreted situation
in the person's awareness

The target situation
in the world

**Fig. 3 - An illustrative example of situation awareness**

According to Devlin [1995], a human agent *individuates* (picks out, or recognizes) objects within a situation and recognizes properties of and relationships among these objects. Accordingly, situation awareness occurs when *an observer recognizes properties of and relations among the interpreted objects through a process of observing a situation in the world.* The observer is a subject in the perceived situation, who observes the situation, or the parts of the situation that can be observed, using sensors of the observer. The observer interprets the observed aspects of the situation to produce an interpreted situation. Fig. 3 shows a perceived situation in which the eye in the figure is a sensor of an observer, the two small vehicles in the target situation are the observed, and the two wheeled vehicles (the left two vehicles in the figure) in the interpreted situation for the observer are the interpreted (or reported) objects.

Predictive Situation Awareness (PSAW) emphasizes the ability to make predictions about aspects of a temporally evolving situation. Decision makers have performed PSAW using higher-level information fusion in which they use the results of low-level fusion to estimate and predict the evolving situation over time. For performing PSAW, a model representing an evolving situation is necessary. In the military domain, a predictive model is defined as the following.

"*Predictive model. A model in which the values of future states can be predicted or are hypothesized; for example, a model that predicts weather patterns based on the current value of temperature, humidity, wind speed, and so on at various locations* [DoD 5000.59-M, 2011]".

A predictive model for PSAW (or a PSAW model) should be flexible enough to represent a variety of complex situations to capture attributes of, relationships among, and processes associated with various kinds of entities in a situation. Also, the PSAW model requires a representation treating uncertainty.

"*Military situations are inherently uncertain, and the available data are inevitably noisy and incomplete. It is essential to be able to represent and reason with uncertainty* [Costa et al., 2009]".

To identify the PSAW model more precisely, we use the following as our working definition of PSAW.

**Definition 3.1 (Predictive Situation Awareness)** Predictive Situation Awareness (PSAW) is the ability to estimate and predict a possible situation involving multiple actors (*Atr*) and/or objects (*Obj*) in different locations (*Loc*), in which actors may trigger events (*Evt*) or activities (*Act*) occurring over time (*T*), and where the meaning of the situation (*MS*) is revealed by integrating previous knowledge with evidence from multiple sources.

Actors (*Atr*) are entities which can generate events (*Evt*) or activities (*Act*), whereas objects (*Obj*) can be involved in events or activities but cannot actively initiate them. For example, vehicles *V1* and *V2* can be modeled as actors, and they may be able to perform *Move*, *Defend*, or *Attack* activities. (In a richer representation, we might model the vehicles as objects and their drivers as actors.) On the other hand, a region or stone is an object and cannot perform an activity. Events and activities happen at a given time or during a time interval (*T*). For example, two vehicles *V1* and *V2* may perform certain activities at time *T1*. The meaning of the situation (*MS*) means why those activities stimulated by the actors are occurring.

The elements (e.g., actor, event, and activity) in PSAW have not been defined universally and various definitions exist. In the followings, we define each element of PSAW used for this research.

**Definition 3.2 (Actor)** An actor (*Atr*) is an entity which can generate events or activities.

**Definition 3.3 (Event)** An event (*Evt*) is something occurring at a given time.

**Definition 3.4 (Activity)** An activity (*Act*) is the something that an actor does during a time interval.

**Definition 3.5 (Meaning of the Situation)** The meaning of a situation (*MS*) is an interpretation or an explanation by an observer of a situation.

**Definition 3.6 (Observer)** An observer (*OR*) is an actor who can be aware of objects through certain senses, interpret the observations, and produce a meaning of a situation (*MS*).

**Definition 3.7 (Sensor)** A sensor (*SR*) is an object which transmits an input event to output event for a certain purpose.

**Definition 3.8 (Target)** A target (*TR*) is something which exists in spatial and temporal spaces and is observed in a target situation by observers.

**Definition 3.9 (Reported Target)** A reported (or interpreted) target (*RT*) is something which is transformed from an observed object by observers and exists in an interpreted situation.

**Definition 3.10 (Observing Condition)** An observing condition (*OC*) is a condition which affects an observing process performed by an observer.

For example, vehicles *V1* and *V2* can be modeled as actors, and they may be able to perform *Move*, *Defend*, or *Attack* activity, and generate various events at time *T1*. If the two vehicles' activities are approaching to a certain target, the meaning of the situation (*MS*) for the two vehicles may be an attack situation which is interpreted by an observer. The meaning of the situation is identical with Devlin's [1995] type of the situation.

The observing process operates in various ways according to the context and type of the observer. The observing condition can be classified into two categories (*Internal Condition* and *External Condition*). The internal condition is associated with the observer (e.g., the observer's observing capability). The external condition concerns the relationship between the observer and the observed objects (e.g., range and weather).

**Definition 3.11 (Target Situation)** The target situation (*TS*) is a situation which contains the relations of the observed objects and is observed by an observer.

**Definition 3.12 (Interpreted Situation)** The interpreted (*IS*) situation is a situation which contains the relations of the interpreted objects and is interpreted by an observer.

To perform PSAW, we may want to answer various questions. (e.g., "How many military vehicles are going to encounter?", "How high is the current level of danger to the Enterprise?", "Where is the target located?", and "What is the enemy doing and why?" [Waltz & Buede, 1986][Costa, 2005][Dorion et al., 2008]). In our research, questions such as the ones listed above are called *PSAW questions*. The purpose of PSAW is to answer some or all of these PSAW questions. Note that Appendix A presents a list of PSAW Questions.

### 3.2. Properties of a PSAW Model

In this section, we introduce some properties of a PSAW model which will be used for constructing a MEBN model for PSAW, called a *PSAW-MTheory*, addressing the PSAW questions in Section 3.1. Generally, when a model is constructed, the following properties for the model can be considered [Sterman, 1991]: (1) A model boundary, (2) exogenous variables, (3) endogenous variables, (4) a time horizon, and (5) a time granularity. The model boundary identifies the exogenous variables and the endogenous variables from the point of view of the user of the model. If the model is a dynamic model in which some states of the model change over time, the time horizon (i.e., the time period the model is representing) and time granularity (i.e., the length of time step) should be identified. These are key modeling properties, which can be used for developing the properties of the PSAW model.

### A) Time and Space

Time and space are basic elements for PSAW. Time can be considered as discrete or continuous. If time is modeled as discrete, time can be represented as a time stamp (e.g., $t_1$, $t_2$, and $t_3$). If time is modeled as continuous, time can be represented as a continuous value with units (e.g., 5.132 seconds). Dynamic Bayesian networks (DBNs) [Murphy, 1998] follow the discrete time model, while Continuous Time Bayesian Networks [Nodelman et al., 2002] use the continuous time model. The continuous time model provides more precise estimation and prediction, however computation of the continuous time model is difficult. In this research, the PSAW-MTheory is based on the discrete time model.

A discrete time in the discrete time model is a time stamp or time slice. A time stamp consists of a starting time and a time period. Time stamps can be equally spaced, which means that the time periods for all time stamps are equal. If time stamps are not equally spaced, then the time periods for all $t_i$ should be described in some way. This approach makes it more complex to treat each time period. Identifying the time period is important and it influences the PSAW model. For example, if a time period for a model is modeled as a long-term (i.e., low precision), it may be difficult to represent events occurring in the short-term with any precision. The time period can be chosen by consulting historical data, expert opinions, and/or sensor capabilities. In this research, we assume a single and fixed time period (e.g., a year, a month, a week, and a day) and leave the problem of a multi-resolution temporal model, in which varying time periods can be associated with each other, for future work [Bettini et al., 1998][Bettini et al., 1998 May]. If there are two more time stamps, various relationships between the time stamps can be held. Allen [1983] proposed thirteen relations; equality($t_i$, $t_j$), six relationships (precedes($t_i$, $t_j$), meets($t_i$, $t_j$), overlaps($t_i$, $t_j$), starts($t_i$, $t_j$), finishes($t_i$, $t_j$), during($t_i$, $t_j$)), and other six relationships with reversed arguments from the first six relationships.

Space can be represented as an *n*-dimensional Euclidean space. A region in space can be a part of Euclidean space. In PSAW, a situation is associated with regions. An object in the situation occupies its own region contained in an overall region of the situation. Randell et al., [1992] defined Region Connection Calculus (RCC) which specified several connections between regions. RCC contains 8 types of connection: (1) DC(x, y) x is disconnected from y, (2) EC(x, y) x is externally connected to y, (3) PO(x, y) x partially overlaps y, (4) EQ(x, y) x is identical with y, (5) TPP(x, y) x is a tangential proper part of y, (6) NTPP(x, y) x is a non-tangential proper part of y, (7) TPPi(x, y) x is a tangential proper part inverse of y, and (8) NTPPi(x, y) x is a non-tangential proper part inverse of y [Randell et al., 1992]. Using RCC, objects in a situation can be represented more clearly and spatial relationships between objects can be reasoned.

Another issue about space and time is whether they are a composite entity or they are mutually incompatible entities. Grenon & Smith [2004] proposed a separation between space and time as named continuant and occurrent, respectively. A continuant continuously exists over time (e.g., objects) and an occurrent changes continuant things over time (e.g., event, processes, and activities). These can be combined into a composite entity (e.g., an object whose shape changes over time).

### B) Object, Actor, and Non-Actor

An object has at least one attribute (e.g., type, size, shape). An object can influence another object (e.g., rain causes the target's speed to be slow) or be related to another object (e.g., the target is behind the tree). Relationships can be static or may change over time (i.e., dynamic relationships). A PSAW model should be able to represent objects, attributes, and static or dynamic relationships (e.g., the target behind the tree at time 1 moved from the tree at time 2).

An object can be classified into two categories (Actor and Non-actor). An actor can engage in an activity or initiate an event. Once the activity or event happens, the actor may have a goal, purpose, mission, plan, preference, and/or belief. These mental attributes of the actor should be considered in the PSAW model. We will discuss this in Section 3.3.

### C) Groups of Objects

In PSAW, a situation is about more than just individual objects. Thus, the situation can contain a group of individual objects. The group can be represented by relationships of the individual objects. For example, countries *A* and *B* may be grouped because they are in an alliance. Many objects can be grouped according to a certain relationship. The relationship combining objects is determined by an observer's perspective. An example is a road on which several types of cars are running. These cars can be grouped by the direction, the speed, or the type. This relationship depends on the intentions of the observer and isn't constrained by space. For example, one person in Asia and another person in Europe can be grouped, if they communicate with each other and/or they work for the same organization. Once some objects are grouped, a property of the group can be observed (e.g., the group's direction or speed). As the single object case, these properties of a group can change over time. Also, the objects in the group can stay or leave the group over time. Furthermore, some groups can be grouped at a higher level. For example, a military company (higher level group) can consist of three platoons (sub-groups).

### D) The observer, the sensor, and the observed

The observer observes the observed through sensors at a certain time. Observations may be intermittent because of observing problems, capabilities of the observer, or concealment by the observed. For this reason, a tracking method estimating an actual property of the observed is necessary. Furthermore, to understand some activities or properties of the observed, the intention or mission of the observed should be identified, so that the observer can estimate the current states of the observed more accurately as well as predict the future states of the observed.

According to the capability of an observer, the observation quality varies. The quality can be regarded as the performance of the observer. This performance should be represented in a PSAW model. At some times, several observers may observe one target and the observers' performances may be different, so the PSAW model should have a means of increasing estimation quality by combining multiple observations. In some cases, one observer can detect several objects and these objects should be distinguished and grouped.

### E) Types of Uncertainty

Uncertainty about a situation can be classified into five categories [Wright et al., 2001][Laskey et al., 2001][D'Ambrosio et al, 2001]: (1) *Existence uncertainty*, (2) *Type uncertainty*, (3) *Reference uncertainty*, (4) *Attribute uncertainty*, and (5) *Relationship uncertainty*. Existence uncertainty is uncertainty about whether an object exists or not. For example, an observation for an object may indicate the actual existence of the object or can be a false alarm. Type uncertainty means uncertainty about what type of object it is. For example, it may be uncertain whether a target is a vehicle or a human. If an object has an attribute that is filled by another entity, then reference uncertainty is uncertainty about which entity fills the slot. An example of reference uncertainty is uncertainty about which object a sensor is looking at. Attribute uncertainty is uncertainty about properties of an object (e.g., the speed of the target). Relationship uncertainty is uncertainty about properties that relate two or more objects (e.g., whether two targets are communicating).

### 3.3. Physical and Mental Situations in PSAW

An observer can observe, interpret, estimate, and predict various situations associated with targets. Commonly, these situations are regarded as physical situations. However, these situations may not be the only physical situations. We can think of what situation of a target which an observer wants to see. For example, a target may act within a decision-making process to achieve certain goals or missions. The decision-making process may influence the actions of the target. The actions may result in some states in the world. And the states may be observed by an observer observing the target. In view of the observer, what the observer wants to see can be anything in the situation of the target (e.g., the goals, the actions, and the states of the target). The situation of the target can be considered in two categories: a *physical* and *mental situation*. If a target is an object (e.g., stone) which doesn't have thinking ability, the physical situation (e.g., a target's size or movement) of the object can be considered alone. If a target is an actor (e.g., human) which can think of something and have an intention, the physical situation of the actor as well as the mental situation of the actor (e.g., a target's emotion or intention) should be dealt with.

For modeling the physical situation, various ontologies for situation have been researched. For reusability of a situation model, a core-ontology for situation awareness [Matheus et al., 2003] was suggested, so that it can be applied to designing various situation models. It provides a schema containing objects, attributes of the objects, and relationships among the objects to support SAW. Situation ontology for hierarchical situation modeling was presented [Yau & Liu, 2006]. Situation ontology consists of context layer for determining elements of context and situation layer for aggregation of contexts or atomic situations. Situational context ontology designed for SAW focuses on a person who is in a context and is involved in a situation [Anagnostopoulos et al., 2007]. The situational context ontology is represented in the Web Ontology Language (OWL-DL) [Baader, 2003] providing useful relations for representation of SAW (e.g., spatial relations, temporal relations, and restriction relations). SAW ontologies [Matheus et al., 2005][Yau & Liu, 2006] were surveyed with an evaluation framework [Baumgartner & Retschitzegger, 2006]. The evaluation framework in the survey measured the SAW ontologies in terms of recommended properties in top-level concept (i.e., object, attribute, relation/role, event, and situation) and specific concepts (i.e., space/time, thematic roles, situation type, and situation object) for SAW. These SAW ontologies can be used for modeling the physical situation.

For modeling a mental situation, we can think of Waltz's [2003] adversary mental process containing the following steps: (1) Firstly, goals and values of the adversary are initiated by stimuli, (2) Motives are initialized from the goals by combining with some beliefs, (3) Intentions proceed from the motives, (4) Alternative plans are led by the intentions, (5)

Chosen plans are decided from the alternative plans, and (6) Action commands following the chosen plans are made. And from the action commands resulted from the entire mental process, actual actions, and state changes in the real world influenced by the actions will appear in the physical domain. However, modeling a mental situation is more complicated than modeling a physical situation as [Waltz, 2003] pointed out that "*real-world adversarial systems are complex and not so easily represented by rigid doctrinal and hierarchical models.*"

For modeling a mental situation, it is necessary to consider what mental processes operate to make decisions by an actor and identify what types of outputs from the mental processes are. In the military domain, the decision-making process has been researched thoroughly, because of the significance of decision-making. Several decision-making models were developed with specific processes. For example, Boyd's [1976, 1987] OODA (*Observe – Orient – Decide – Act*), Lawson's [1981] C2 (Command Control) model (*Sense – Process – Compare – Decide – Act*), Buettner's [1985] HEAT (Headquarters Effectiveness Assessment Tool) model (*Monitor – Understand – Develop Alternatives – Predict – Decide – Direct*), Chin et al.'s [1999] decision-ladder (*Activation – Observe data– Identify state – Predict consequences – Evaluate options – Choose tasks – Choose procedure – Carry out activity*), Joslyn & Rocha's [2000] SHOR (*Stimulus – Hypothesis – Option – Response*), US Army's [2012] the Military Decision-Making Process, etc. In this research, we employ OODA as a general mental process and introduce in the next section.

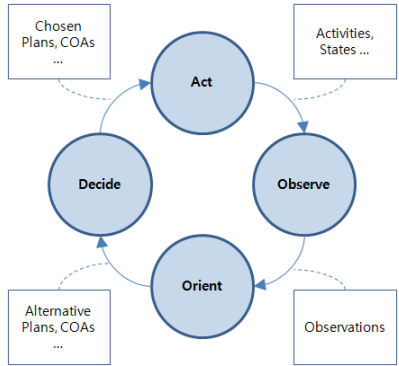

**Fig. 4 - OODA Loop**

### 3.4. Properties of OODA

OODA is a high-level concept for decision making, so OODA can be a starting point to consider more specific processes (e.g., C2 model, HEAT, decision-ladder, SHOR, Military Decision-Making Process, and etc). For example, Grant & Kooter [2005] tried to reengineer OODA into a general C2 process by addressing some deficiencies of OODA (e.g., planning process and learning process). They concluded that OODA could represent a general C2 process. In this research, we employ OODA (Fig. 4) as a general mental process, in which an actor may make a decision. According to the situation in a specific domain, the mental process in our research based on OODA may be replaced by the specific processes.

OODA contains four steps (Observe, Orient, Decide, and Act). Each of the four steps in OODA can have various inputs and outputs as shown Table 1. For example, in the Observe step, data or signal from every mental/physical situation (e.g., states, activities, and goals) of external systems (e.g., an adversary) as well as internal systems (e.g., a command center or an allied army) in the world are observed in some internal observing guidance or control, and observations derived from data or signal are produced. In the Orient step, observations become information, formed as a model, by reasoning, analysis, and synthesis influenced from knowledge, belief, condition, etc. The Orient step can produce plan and COA (Course Of Actions). Hypotheses or alternatives for models can be decided by the preference of a decision maker in the Decide step. In the Act step, all decided results are implemented, and real activities and states can be operated and produced, respectively. The four steps continue until the end of the life cycle of the actor.

Fig. 4 shows a situation in which only one actor operates. We can extend the situation into a multi-actor situation in which many actors interact with each other. The multi-actor situation can be represented as multiple interacting OODAs.

**Table 1. Possible inputs/outputs for the four steps in OODA**

|  | Input | Output |
|---|---|---|
| Observe | World Data, Observing Condition, Internal Guidance/Control | Observations |
| Orient | Knowledge, Belief, Strategies, Experience, Orienting Condition, Observations | Alternative Plans, Alternative COAs, Alternative Responses, Hypotheses |
| Decide | Vision, Mission, Core Values, Goals, Objectives, Intension, Deciding Condition, Alternative Plans, Alternative COAs, Alternative Responses, Hypotheses | Chosen Plans, Chosen COAs, Chosen Responses, Chosen Hypotheses |
| Act | Internal Guidance/Control, Acting Condition, Chosen Plans, Chosen COAs, Chosen Responses, Chosen Hypotheses | Activities in World, States in World |

Fig. 5 shows an example of a two-actor situation with two-OODAs in which the observer observes all steps in OODA of the observed. In this research, we suppose that every step in OODA of the observed can be observed by an observer performing the Observe step in OODA. The Observe step of an actor $i$ at time $t$ can be defined as a set of four steps of another actor $j$, Observe$_{i,t}$ = {Observe$_{j,t}$, Orient$_{j,t}$, Decide$_{j,t}$, Act$_{j,t}$}.

For example, the observer $A$ can see the plan and COA of the observed $B$ as well as the activities and status of the observed $B$ at time 1, Observe$_{A,1}$ = {Observe$_{B,1}$, Orient$_{B,1}$, Decide$_{B,1}$, Act$_{B,1}$}. Furthermore, we can consider that the observed $B$ also can see the observer $A$'s all steps in OODA at the same time 1, Observe$_{B,1}$ = {Observe$_{A,1}$, Orient$_{A,1}$, Decide$_{A,1}$, Act$_{A,1}$}. And, the observer $A$ can see a situation in which the observed $B$ are seeing the observer $A$ (e.g., in a chess game, a player is pondering about the thought of the opponent about which the opponent is observing the player), Observe$_{A,1}$ = {Observe$_{B,1}$, Orient$_{B,1}$, Decide$_{B,1}$, Act$_{B,1}$}, where Observe$_{B,1}$ = {Observe$_{A,1}$, Orient$_{A,1}$, Decide$_{A,1}$, Act$_{A,1}$}. These recursive observations between the observer *A* and the observed *B* (e.g., looking an object in a mirror with another mirror in opposite side of the mirror) can be infinite. However, because of a computational limitation of human or machine, the recursive observations between both will stop at a certain point.

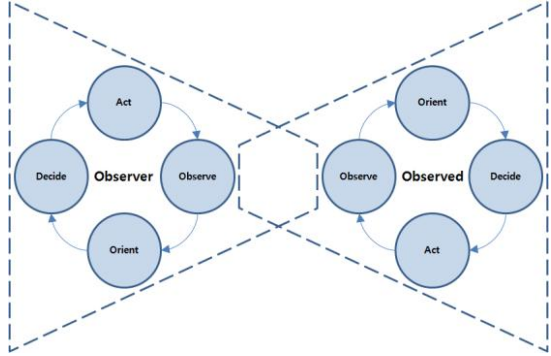

**Fig. 5 - OODA Loop**

In the multi-OODAs, we should consider some properties for the observed and time. Obviously, the observer *A* can observe itself, *i = j*. For example, we can meditate ourselves. In some cases, the observed *B* may not be only one actor, but it is a set of actors, $i = \{a_1, a_2, ..., a_n\}$, where $a_k$ is a unit actor for the set. The time, *t*, when the observer *A* observes the observed *B* can be a discrete time with a certain time period (e.g., 1 sec or 1 hour) or a continuous time.

### 3.5. PSAW in terms of OODA

A PSAW model may need to represent the four steps in OODA and their inputs/outputs to express various operating aspects for an actor. For example, an observer may observe the inputs and outputs of the OODA process of the observed, estimate current situations, and predict the future situation for the observed. These observing, estimating, and predicting operations of PSAW correspond to the Observe and Orient steps in OODA. In other words, the observer in the Observe and Orient steps (or PSAW operations) can observe any of inputs/outputs (in Table 1) of the Observe, Orient, Decide, and Act steps of the observed.

In the view of an observer, all inputs/outputs in the four steps can be considered as two elements: interpreted elements and actual elements. If the elements are comprehended subjectively, they are interpreted elements (i.e., the two wheeled vehicles in the interpreted situation in Fig. 3.) which are derived from the actual elements (i.e., the wheeled and tracked vehicles in the actual situation in Fig. 3.). In other words, the real type of a target (i.e., actual elements) can be different from the reported type of the target (i.e., interpreted elements).

Fig. 6 shows a relationship between the interpreted elements in OODA, called an *interpreted OODA*, and actual elements in OODA, called an *actual OODA*. The figure depicts a situation when the observer observes the observed. In our PSAW model, the observer may see all elements of OODA sequence of the observed. For example, a spy investigates the report, organization, strategy, intention, and activities of the spy's opponent. The first OODA sequence in Fig. 6 is an OODA sequence for the observer who performs the OODA processes from the Observe step to the Act step. The observer in the Observe step can detect any element in the interpreted OODA (i.e., the second OODA) of the observed. The detected (or reported) element in the interpreted OODA sequence is used to estimate the actual current element and predict the future element in the actual OODA (i.e., the third OODA) by the observer in the Orient step.

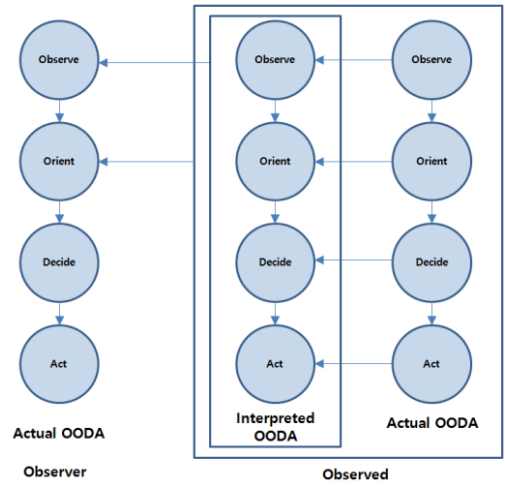

**Fig. 6 - A situation of an observer in terms of Interpreted/Actual OODA**

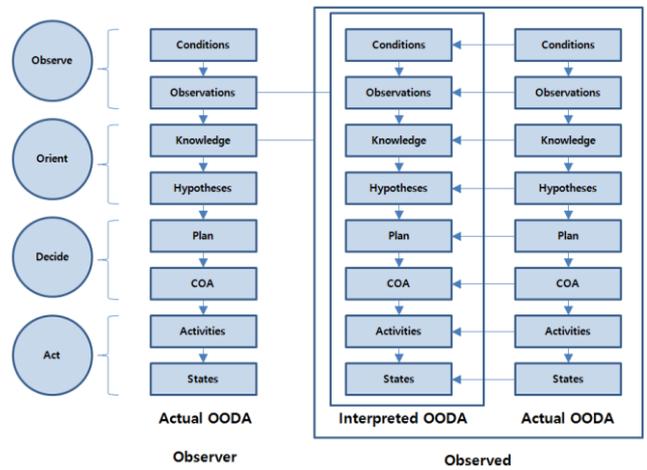

**Fig. 7 - Illustrative example of a PSAW model**

Fig. 7 shows an illustrative example of the above situation with specific elements of OODA. According to requirements of constructing a PSAW model, elements of OODA are chosen. The elements of OODA (e.g., condition, observations, and plan in Table 1) in Fig. 7 can be chosen in terms of the modeling properties (i.e., a model boundary, exogenous variables, endogenous variables, a time horizon, and a time granularity). An observer contains 8 elements of the actual OODA (i.e., observing conditions, observations, knowledge, hypothesis, plan, COA, activity, and states). The observing conditions are a set of conditions when the observer observes the observed (e.g., weather condition). The observations are all elements of the interpreted OODA in the observed. The knowledge in this example model is estimation of all elements of OODAs in the observed. Using the knowledge, the observer estimates possible current situations or predicts future situations. Then the observer decides plans influencing the development of COAs. The observer (now it is an actor following the COAs) executes some activities changing states of the observer, the observed, or the world.

In the next section, we introduce a PSAW-MEBN reference model developed from the PSAW definitions and consideration in this section.

## 4. PSAW-MEBN Reference Model

A PSAW-MEBN reference model specifies reference MFrags, RVs, and entities which support the design of a MEBN model for PSAW. Such a MEBN model is called a *PSAW-MTheory*. The PSAW-MTheory is designed to reason about the PSAW questions from Section 3.1 and can be supported by the reference model, called a *PSAW-MEBN reference model*. The PSAW-MEBN reference model is based on our discussion in the previous sections and provides a common semantics for a PSAW-MTheory, which can be used for different applications. Constructing an MTheory can be depending on various purposes as discussed in [Costa, 2005].

"*Ultimately, the approach to be taken when building an MTheory will depend on many factors, including the model's purpose, the background and preferences of the model's stakeholders, the need to interface with external systems. etc.* [Costa, 2005] ."

The reference model should be useful to construct a PSAW-MTheory in a new application, so it should enable a PSAW-MTheory modeler to construct the PSAW-MTheory efficiently and effectively. Therefore, the reference model should be designed sufficiently to define precisely a PSAW-MTheory and flexibly to be used for various applications. We develop the PSAW-MEBN reference model to satisfy these two criteria (i.e., *sufficient detail* and *flexibility*) by adopting both of the use of expert knowledge and the design concept of modularity.

To make the reference model flexible and in sufficient detail regarding the characteristics of PSAW, the PSAW-MEBN reference model is designed with the following issues in which a PSAW-MTheory modeler may have difficulty. (1) How does the MTheory address the questions of PSAW? (2) What type of entities and RVs can be chosen to represent hypotheses for PSAW? (3) How can MFrags be decomposed for efficiently exploring the hypotheses? (4) What should be the relationships between RVs in the MFrags for to support efficient and accurate inference?

Fig. 8 depicts an illustrative example of an SSBN representing PSAW. This very simplified SSBN example has are five levels: *Situation Identification*, *Mission Identification*, *Activity Identification*, *Object Identification*, and *Detection* (note that these levels are not necessarily general).

At the detection level, there are two types of RVs; a *Sensor Type* and

*Report* RV. The *Sensor Type* RV describes the type of a sensor, *s*, which is detecting a target, *g*, at time *t*, and is producing a report, *r*. The *SensorType_g1_s1_t1* RV means the type of the sensor *s1* detecting the target *g1* at time *t1*. The *Report_r1_g1_s3_t1* RV means the report *r1* of the target *g1* detected by the sensor *s3* at time *t1*.

At the object identification level, the reports of the target from the sensors at the detection level are integrated into an *ObjectState* RV. At the object identification level, there are three *ObjectState* RVs. The first is for the target *g1* at time *t1*. The second is for the target *g1* at time *t2*. The third is for the target *g1* at time *t3*.

At the activity identification level, from the *ObjectState* RVs at times 1~3 in the object identification level, an *Activity* RV for the target *g1* is estimated. Note that the *Activity* RV occurs during times 1~3. This duration is specified by *t123* in the *Activity_g1_t123* RV. As another example, at the activity identification level, the *Activity_g1_t67* RV means the activity runs during times 6 and 7.

At the mission identification level, a *Mission* RV is integrated from the *Activity* RVs at the activity identification level. The *Mission_g1_t1234567* RV represents the mission of the target *g1* during times 1 through 7 covering the duration of the activities belonging to the mission.

At the situation identification level, two *Situation* RVs with distinct time durations (i.e., times 1~9 and times 10~13) are depicted. Some or all *Mission* RVs can influence a *Situation* RV. For example, the RV *Situation_g123_t123456789* is influenced by the *Mission_g1_t1234567*, *Mission_g2_t1234567*, and *Mission_g3_t123456789* RV.

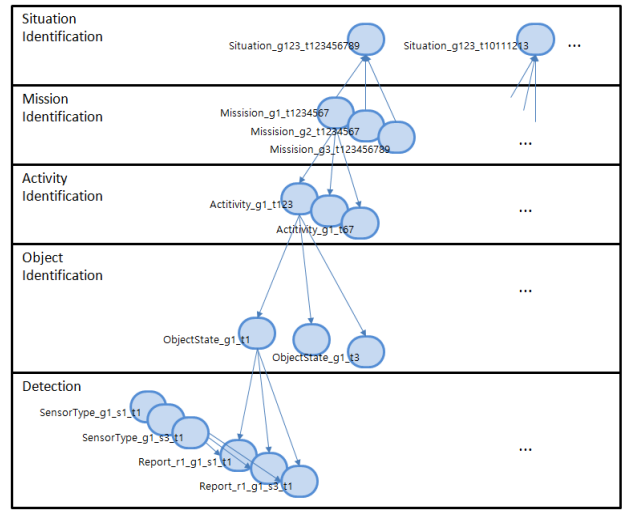

**Fig. 8 - An example of an SSBN generated by a PSAW-MTheory [Park et al., 2014]**

In Section 3.2 and 3.3, We discussed several ontologies such as Space [Randell et al., 1992], Time [Allen, 1983], Situation [Yau & Liu, 2006][Baader, 2003], and SAW [Matheus et al., 2003][Anagnostopoulos et al., 2007]. These ontologies can be used for the development of an SSBN or an MTheory for PSAW. For example, the time relation $precedes(t_i, t_j)$ [Allen, 1983] can be used for an RV representing whether a time $t_i$ finishes before a time $t_j$ starts. The space relation EC($x$, $y$) [Randell et al., 1992], $x$ is externally connected to $y$, can be used for an RV representing whether two regions $x$ and $y$ adjoin each other and they do not overlap. However, although in a simple case (e.g., two entities are associated), using such a relationship may be easy, it becomes quite complicated in situations where more than two entities are associated. Also, combining probabilistic information about multiple overlapping relationships may create very challenging issues for representation and reasoning. In this research, we focus on the simple case and leave such complicated situations as a future research topic.

The five levels of Fig. 8 can address the PSAW questions from Section 3.1. For example, the object identification level can answer questions such as "what type is the target?". The activity identification level can answer questions such as "what are the next activities performed by the (grouped) targets?". The mission identification level can answer questions such as "what is the meaning of the situation?". The situation identification level can answer questions such as "how much danger is there?".

The SSBN example illustrates how an SSBN for PSAW can be composed. In the following sections, we discuss how an MTheory for PSAW can be composed. In Section 4.1, special entity types for the reference model are presented. In Section 4.2, RV types based on the entity types are presented. In Section 4.3, MFrags for the reference model are discussed.

### 4.1. Entities for the Reference Model

The examples in Fig. 3 and Fig. 8 allow us to identify some fundamental elements for PSAW. In this research, the observer, the sensor, the target, the reported target, and the time are determined as five fundamental elements. These elements can be represented as entities in a PSAW-MTheory. In PSAW, a target is meaningful, when it can be identified in terms of what it is and what it does. An activity operated by nothing or non-target is meaningless. And also a target which just exists without any operation (or changes) forever is meaningless. In PSAW, we presume that there is always an observer observing (or being related to) the target through sensors over/at the time. If these fundamental elements don't exist, we can't be aware of a situation.

In MEBN, an entity type is a unique kind of thing, and can be instantiated to one or more entity instances which exist distinctly and independently. For example, *Person* can represent an entity type instantiating person entity instances (e.g., *John* and *Mathew*). An entity in the reference model can be classified into five categories, called *PSAW-Entities*; (1) the time entity $T$, (2) the observer entity *OR*, (3) the sensor entity *SR*, (4) the reported target entity *RT*, and (5) the (actual) target entity *TR*. Entities derived from these categories describe a situation in which an observer observes a target through sensors and interprets it as a reported target at a certain time. For example, an omnidirectional radar (as *SR*) detects a vehicle object (as *TR*) and produces a vehicle object report (as *RT*) at time *T1* (as *T*). A user who uses a PSAW-MTheory can be represented as the observer entity *OR*. Commonly, it is not necessary to represent the observer entity explicitly in a PSAW-MTheory, because we already know about who are using the PSAW-MTheory. However, there are special situations in which a relationship between a target and an observer should be represented in a PSAW-MTheory (e.g., a target takes aim at an observer).

### 4.2. Random Variables for the Reference Model

A Random Variable (RV) represents the uncertainty of a set of values in mutually exclusive (i.e., events cannot occur simultaneously) and collectively exhaustive (i.e., a set of events covers all sample space outcomes). In MEBN, random variables (RVs) (e.g., resident, input, and context nodes) can contain ordinary variables (written as strings of lowercase letters) that can be filled by instances of entities (written as strings of uppercase letters). For example, *Isa* context nodes are used to specify the types of the entities that can fill an ordinary variable. Isa($p$, *PERSON*) is an *Isa* context node. Only entities of type *Person* can fill in for $p$ when creating instances of RVs in an MFrag containing the Isa($p$, *PERSON*) node.

In the reference model, random variables containing ordinary variables from entity types $T$, *OR*, *SR*, *RT*, and/or *TR* are called *PSAW-RVs*. The reference model doesn't limit the number of each of the ordinary variable types in PSAW-RVs. For example, a situation in which two vehicles communicate with each other and are detected by a COMINT sensor at time *T1* can be described by a PSAW-RV, specified by Communicated($v1$, $v2$, $cmt$, $t$), where $v1$, $v2$, $cmt$, and $t$ are ordinary variables that can be filled in by a first vehicle entity (*TR*), a second vehicle entity (*TR*), a COMINT sensor entity (*SR*), and a time entity (*T*), respectively. Thus, two target entities (*TR*) are used in the PSAW-RV.

Each of the five entity types in Section 4.1 can have its own attributes. For example, an attribute of the time entity can be used for indicating the clock time associated with a timestamp variable, RealTime($t$) (e.g., RealTime($t1$) is 12:10:01 UTC for the time entity $t1$). An attribute for the sensor entity can represent the type of the sensor, SensorType($sr$) (e.g., SensorType($sr1$) can be an omnidirectional radar). Entities of the five entity types can be related to each other (e.g., SensorCondition($sr1$, $t1$) and IsAlliance($or1$, $tr1$)).

Semantically, the type for a PSAW-RV can depend on the context where an observer, sensor, target, target report, and times are involved. For example, Little & Rogova [2005] suggested main relations for continuant and occurrent things: (1) Topology/mereology relations (e.g., disjoint, joint, overlap, cover, reachable, unreachable, contain, and part of), (2) Direction (e.g., along, towards, east, west, south, north, similar, and opposite), (3) Distance (e.g., far, very far, near, and very near), Size (e.g., small, large, and same), (4) Relationships between time points (e.g., before, at the same time, start, finish, soon, very soon, resulting in, and initiating), and (5) Relationships between time intervals (e.g., disjoint, joint, overlap, inside, and equal). To develop a PSAW-RV, these relations which represent physical aspects can be considered. For the mental situations (Section 3. 3), the elements of OODA in Table 1 can be used. As the elements of OODA in Fig. 7, the observing conditions, observations, knowledge, hypothesis, plan, COA, activity, and states can be the PSAW-RVs.

These PSAW-RVs can be classified into four kinds of RV: Observing condition RV (*OC_RV*), Reported object RV (*RT_RV*), Target object RV (*TR_RV*), and Situation RV (*SIT_RV*). Fig. 9 shows these core kinds of RV (*OC_RV*, *RT_RV*, *TR_RV*, and *SIT_RV*) and their instances.

**Definition 4.1 (Observing Condition RV)** An observing condition RV, *OC_RV*, is an RV indicating a condition of an observer or a sensor. An observing condition RV can depend on other observing condition RVs, $OC\_RV_I \rightarrow OC\_RV_j$, where *I* is a set of indices for *OC_RVs* and $j \notin I$.[‡]

The observing condition RV can be decomposed into two sub-types (An *internal condition RV* and an *external condition RV*). An internal condition RV is an RV used for a condition of the internal states of the observer or the sensor. For example, a capability of an omnidirectional radar *sr1* at time *t1* can be represented as the external condition RV Capability(*sr1*, *t1*). An external condition RV is an RV used for an observing condition between the observer/sensor and the observed. For example, the range between an omnidirectional radar *sr1* and a moving target *t1* at time *t1* can be represented as the internal condition RV Range(*sr1*, *tr1*, *t1*).

**Definition 4.2 (Reported Object RV)** A reported (or interpreted) object RV, *RT_RV*, is an RV used for a reported object from the observer. A reported object RV can depend on a target object RV and/or an observing condition RVs, $TR\_RV \rightarrow RT\_RV$ and/or $OC\_RVs \rightarrow RT\_RV$.

For example, if a speed of the moving target *tr1* observed by the omnidirectional radar *sr1* at time *t1* is interpreted as a reported speed of a reported moving target *rt1*, this situation can be represented as the reported object RV ReportedSpeed(*rt1*, *t1*) depending on the target object RV Speed(*tr1*, *t1*) and the observing condition RV Range(*sr1*, *tr1*, *t1*). The observations for targets can be reported object RVs

**Definition 4.3 (Target Object RV)** A target object RV, *TR_RV*, is an RV used for actual attributes and/or relations for the target object. A target object RV can depend on other target object RVs, $TR\_RV_I \rightarrow TR\_RV_j$, and/or situation RVs, $SIT\_RV_s \rightarrow TR\_RV_j$, where *I* is a set of indices for *TR_RVs* and $j \notin I$.

For example, an actual speed Speed(*tr1*, *t1*) of the actual moving target *tr1* at time *t1* can depend on the type of the actual moving target *tr1* TargetType(*tr1*). The states, activities, plans, and missions of targets can be target object RVs.

**Definition 4.4 (Situation RV)** A situation RV, *SIT_RV*, is an RV representing an aspect of a situation. The situation RV can depend on target object RVs, $TR\_RVs \rightarrow SIT\_RV_i$, and/or other situation RVs, $SIT\_RV_I \rightarrow SIT\_RV_j$, where *I* is a set of indices for *SIT_RVs* and $j \notin I$.

For example, if there is a situation in which two moving targets (*tr1*, *tr2*) are attacking at time *t1*, this situation can be represented as an RV AttackingSituation(*tr1*, *tr2*, *t1*) which depends on target type RVs TargetType (*tr1*) and TargetType (*tr2*).

Determining causality between PSAW-RVs is a difficult task because there may be possible unknown or hidden mechanisms influencing the events. The task may rely on experts' knowledge and/or machine learning. Nevertheless, some restrictions to determine causality between events can be considered. We can assume that between two events (i.e., *A* and *B*) if the event *A* occurs before the event *B*, then the event *B* can't cause the event *A*. For example, a PSAW-RV Communicated(*v1*, *v2*, *cmt*, *t2*) can't cause to a PSAW-RV Communicated(*v1*, *v2*, *cmt*, *t1*), but the opposite may be possible. Fig. 9 shows possible causal relationships. The observing condition RVs and target RVs can cause the reported object RVs. The target RV can cause the situation relation RVs. Fig. 9 contains three layers (*Situation*, *Object*, and *Observation*). The observation layer and the situation layer in Fig. 9 correspond to the detection and the situation identification in Fig. 8, respectively. The object layer in Fig. 9 contains the mission identification, the activity identification, and the mission identification object in Fig. 8.

---

‡ A → B means B depends on A.

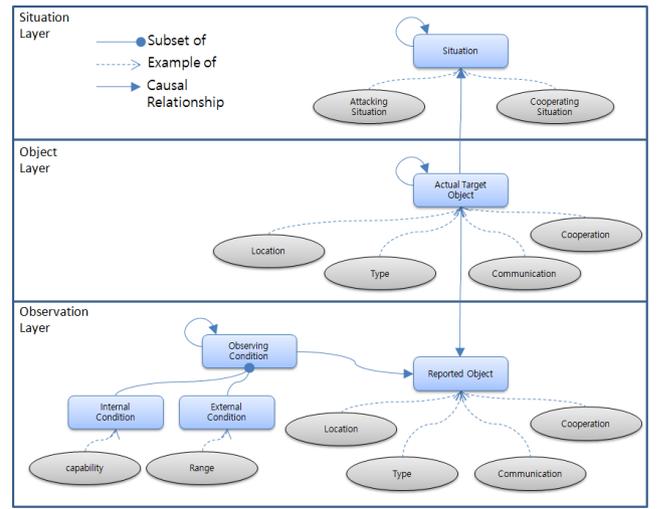

**Fig. 9 - Four core kinds of RVs and examples of each**

### 4.3. The Five MFrag Groups

In MEBN, an MFrag is a fragment of a probabilistic graphical model in which the nodes contain variables that are placeholders for domain entities. MFrags can be instantiated with domain entities and combined into complex models with repeated structure. MFrags for PSAW, called *PSAW-MFrags*, contain PSAW-RVs as defined in Section 4.2. The PSAW-MEBN reference model suggests five groups: (1) Context, (2) Target Object, (3) Observing Condition, (4) Reported Object, and (5) Situation. These five groups can be constructed as five types of MFrags: (1) a *Context* MFrag, (2) a *Target Object* MFrag, (3) an *Observing Condition* MFrag, and (4) a *Reported Object* MFrag, and (5) a *Situation* MFrag, respectively. Fig. 10 shows an example of PSAW-MFrags for an illustrative PSAW domain.

#### A) Context MFrag

MFrags in this group define distributions for resident nodes which are used for context nodes in other MFrags. To construct some PSAW-MFrags (e.g., *Target*, *Observing Conditions*, *Report*, and *Situation*), contexts for each MFrag may or may not be defined. For example, the *observing conditions* MFrag has a context node ObserverOf(*sr, tr*) to determine a sensor entity and a target entity, where the sensor entity is an observer of the target entity. For context MFrags, three common context nodes (*Predecessor*, *ObserverOf*, and *ActualObject*) can be considered. We will discuss these RVs in the following MFrags.

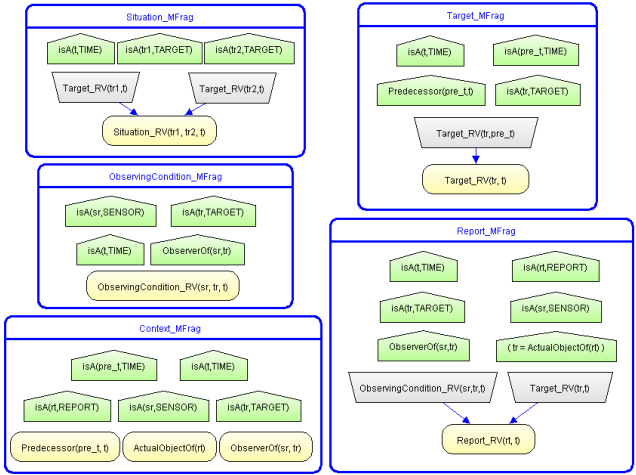

**Fig. 10 - Illustrative example of PSAW-MFrags**

### B) Target MFrag

MFrags in this group define distributions for PSAW-RVs related to relations and attributes of targets. This MFrag contains a resident node *Target RV* and an input node *Target RV*, a recursive relationship. These nodes can have target identifiers and time identifiers. The difference between the resident node and the input node is that the input node has a previous time identifier and the resident node has a current time identifier. Therefore, the resident node at the current time is influenced by the input node at the previous time. To determine this time relationship, the context node *Predecessor(pre_t, t)* is included in the MFrag. The node *Predecessor(pre_t, t)* means that the time interval *pre_t* occurs immediately before the time interval *t*. The node *Predecessor* has its value set deterministically to represent sequential time steps.

This MFrag illustrates possible resident nodes and possible input nodes related to relations/attributes of a target. In a specific usage, a wide variety of nodes is possible. If the relations/attributes of the target do not change over time, the time identifiers in the resident nodes are omitted, and the input nodes for the resident nodes and the context node *Predecessor* are not necessary.

### C) Observing Conditions MFrag

This MFrag represents knowledge about conditions of the sensor. This MFrag contains a resident node *Observing Condition RV* with the sensor identifiers, the target identifiers, and the time identifiers as its possible values. To determine the relationship between the sensor identifiers and the target identifiers, the context node *ObserverOf(sr, tr)* is included in the MFrag. The node *ObserverOf(sr, tr)* means that a sensor *sr* is the observer of a target *tr*, and has its value set deterministically. The context node *ObserverOf* allows a situation in which a sensor observes many targets and a target is observed by many sensors as well.

### D) Report MFrag

This MFrag represents knowledge about relations of a report object. This MFrag contains a resident node *Report RV* and two input nodes *Observing Condition RV* from *Observing Condition MFrag* and *Target RV* from *Target MFrag*. The node *Report RV* has the report identifiers and the time identifiers as its possible values. To determine the relationship between the target and the report object, the context node *tr = ActualObject(rt)* is included in the MFrag. The context node *tr = ActualObject(rt)* means that if a target *tr* is the actual object of a reported target *rt*, the context node is true. This context node makes a relationship between a reported object (e.g., a reported type from a report) and an actual target (e.g., an actual type from an actual target).

### E) Situation MFrag

This MFrag defines distributions for PSAW-RVs related to a situation for targets. In the example, the *Situation* MFrag contains a resident node *Situation* and two *Target RV* input nodes for each target. The resident node *Situation* can have the target identifiers and the time identifiers. In this MFrag, only two targets are treated, however many targets are designed in such a *Situation MFrag*. This MFrag integrates all or some relations between targets to produce a new aspect from these relations. The situation can vary according to what the observer related to this MFrag want to see. For example, the situation can be whether two vehicles are in the same group, whether two vehicles are communicating to each other, what a relationship between two vehicles is, what causality between two vehicles exist, and how activities from two vehicles are related to each other.

In PSAW, understanding a situation in which targets operate for their own purposes is one of the important issues. Identifying just the type of a target is an insufficient task for PSAW. The meaning of *awareness* is not to perceive and estimate actual properties of a target, but is to understand, interpret, and explain the relationships between targets. Kokar et al [2009] stated: "*The main part of being aware is to be able to answer the question of "what's going on?"*". Awareness of a situation is subjective according to an observer, who is aware of the situation. The modeler, who is developing a probabilistic ontology to support PSAW, should define what situation will be considered and explained through all observation from the world.

### 4.4. Summary of the PSAW-MEBN Reference Model

Table 2 summarizes the overall elements for the PSAW-MEBN reference model. The PSAW-MEBN reference model contains three groups: (1) PSAW-Entities (e.g., *Time Entity T*), (2) PSAW-RVs (e.g., *Observing Condition OC_RV*), and (3) PSAW-MFrags (e.g. *Context MFrag*). PSAW-RVs contain the definitions of the causal relationships (e.g., Fig. 9). For example, in the table, {*OC_RV*, *TR_RV*} → *RT_RV* means that the reported object *RT_RV* depends on the observing condition *OC_RV* and target object *TR_RV*.

These elements can be exploited, modified, and extended according to a specific PSAW domain (e.g., PSAW for a smart farm, a smart manufacturing, a smart health, a smart city, and a smart government). In the following section, we introduce two applications of the PSAW-MEBN reference model.

**Table 2. Summary of the PSAW-MEBN Reference Model**

| Name | Elements |
|---|---|
| *PSAW-Entities* | *Time Entity T* <br> *Observer Entity OR* <br> *Sensor Entity SR* <br> *Reported Target Entity RT* <br> *Target Entity TR* |
| *PSAW-RVs* | *Observing Condition OC_RV* <br> *Reported Object RT_RV* <br> *Target Object TR_RV* <br> *Situation SIT_RV* |
| *Causal Relationships* <br> *between PSAW-RVs* | $\{OC\_RV\} \rightarrow OC\_RV$ <br> $\{OC\_RV,\ TR\_RV\} \rightarrow RT\_RV$ <br> $\{TR\_RV,\ SIT\_RV\} \rightarrow TR\_RV$ <br> $\{SIT\_RV,\ TR\_RV\} \rightarrow SIT\_RV$ |
| *PSAW-MFrags* | *Context MFrag* <br> *Observing Condition MFrag* <br> *Report MFrag* <br> *Target MFrag* <br> *Situation MFrag* |

## 5. Use Case

In this section, we introduce two example use cases using the PSAW-MEBN reference model to develop PSAW-MTheories for two proof-of-concept PSAW systems: a Smart Manufacturing System and a Maritime Domain Awareness System.

### 5.1. Smart Manufacturing System

Manufacturing is the process in which manufacturers design, develop, produce, and provide products to markets. Smart manufacturing uses the latest technologies (e.g., Artificial Intelligence (AI), Cloud Computing, Internet of Things (IoT), Cyber-Physical Systems (CPS), and Big Data) to respond proactively, responsively, and adaptively to market requirements. Smart manufacturing requires capabilities such as self-awareness, self-prediction, self-reconfiguration, and self-maintenance. The initial step to support smart manufacturing can be to perform Predictive Situation Awareness (Section 3). We call such PSAW for manufacturing as MSAW.

Park et al. [2017] introduced a PSAW model for smart manufacturing using MEBN, called an MSAW-MEBN model. The MSAW-MEBN model aims to estimate current situations as well as to predict future situations in manufacturing. The MSAW-MEBN model was developed via the PSAW-MEBN reference model (Section 4), to reason about complex and uncertain situations in manufacturing. In the following, the MSAW-MEBN model is introduced.

#### A) MSAW-MEBN Model developed from the PSAW-MEBN Reference Model

A decision maker in charge of management for manufacturing may need some overall situation information (e.g., manufacturing states, quality of products, total manufacturing cost, and total production time). The PSAW-MEBN reference model (Section 4) provides some information on how to represent such situations. The MSAW-MEBN model was derived using the PSAW-MEBN reference model. Fig. 11 shows six core RVs and their causal relationships in the MSAW-MEBN model, and also includes examples of the core RVs.

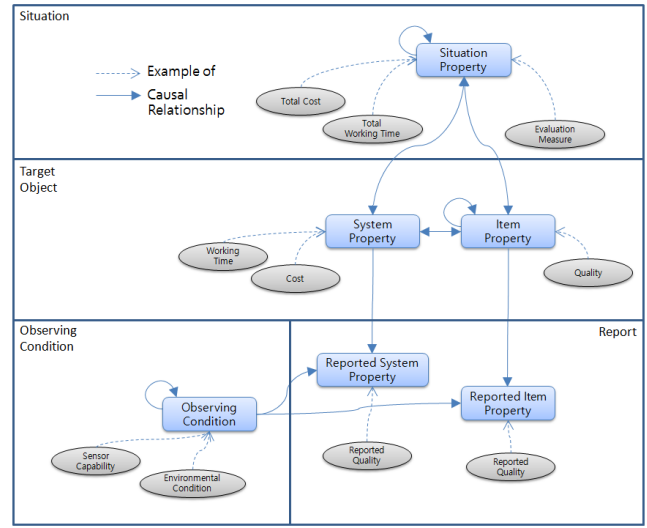

**Fig. 11 – Six core RVs and their causal relationships in the MSAW-MEBN model derived using the PSAW-MEBN reference model [Park et al., 2017]**

In the MSAW-MEBN model, the (actual) target object RV in the PSAW-MEBN reference model is divided into two RVs: *System Property* RV and *Item Property* RV. A system that performs processes in manufacturing produces outputs (e.g., products and materials) by receiving inputs (e.g., energy and resources). Therefore, the basic unit of manufacturing can be defined as systems, inputs, and outputs. The inputs and outputs, commonly called items, are defined as entities with similar properties. We consider the systems and items (i.e., inputs and outputs) to be the basic elements of manufacturing.

Each system has its own properties (or attributes), and each item also has properties. Examples of system properties include operating time and production cost. An example of an item property is quality. In the MSAW-MEBN model, these properties are considered as RVs (e.g., system property RV and item property RV).

System properties and item properties can be influenced by each other. Item properties can affect other item properties. For example, in a steel plate manufacturing factory, a heater is used to preheat slabs before rolling. Heating cost and heating time are properties of the heater (i.e., System). Slab temperature and slab size are properties of items. The working time of the heater may depend on the size of the slab. Thus, the input item property (i.e., the slab size) influences the system property (i.e., the heating time). In addition, the slab temperature before heating affects the slab temperature after heating. Thus, the input item property influences the output item property. These causal relationships are shown in Fig. 11.

Table 3 summarizes the elements for the MSAW-MEBN model extended from the PSAW-MEBN reference model.

**Table 3. Summary of the MSAW-MEBN Model [Park et al., 2017]**

| Name | Elements |
|---|---|
| MSAW-Entities | Time Entity T<br>Observer Entity OR<br>Sensor Entity SR<br>Reported Target System Entity RS<br>Reported Target Item Entity RI<br>Target System Entity TS<br>Target Item Entity TI |
| MSAW-RVs | Observing Condition OC_RV<br>Reported System Property RSYS_RV<br>Reported Item Property RIT_RV<br>System Property SYS_RV<br>Item Property IT_RV<br>Situation Property SIT_RV |
| Causal Relationships between MSAW-RVs | {OC_RV} → OC_RV<br>{SYS_RV, OC_RV} → RSYS_RV<br>{IT_RV, OC_RV} → RIT_RV<br>{IT_RV} → SYS_RV<br>{SYS_RV, IT_RV} → IT_RV<br>{SYS_RV, IT_RV} → SIT_RV |
| MSAW-MFrags | Context MFrag<br>Observing Condition MFrag<br>Report MFrag<br>Target (System, Input, Output) MFrag<br>Situation MFrag |

Note that the MSAW-MEBN model can contain five MFrags discussed in the PSAW-MEBN reference model in Section 4. In the MSAW-MEBN model, the *Target* MFrag is divided into three MFrags: *System*, *Input*, and *Output* to represent the manufacturing process.

### B) Discussion

In this research, we introduced an MSAW-MEBN model, developed via the PSAW-MEBN reference model (Section 4), to reason about complex and complicated situations in manufacturing. A use case in which the MSAW-MEBN model was applied can be found in [Park et al., 2017]. The MSAW-MEBN model in the use case supports PSAW for a steel plate manufacturing process to predict the total cost, total time, and total quality rate. In this research, we were able to develop the MSAW-MEBN model using the PSAW-MEBN reference model in a short time (7 days) in order to support the development of a smart manufacturing system.

### 5.2. Maritime Domain Awareness System

PROGNOS [Costa et al., 2009] is a proof-of-concept system to support Maritime Domain Awareness (MDA). The existing system for MDA (e.g., US Navy's Net-Centric infrastructure, FORCENet) is used to fuse lower-level multi-sensor data, analyze the fused data by human analysts, and support decision-making for naval operations. However, the era of big data requires greater automation. PROGNOS supports ingestion of lower-level data, fusion of heterogeneous input, and probabilistic assessment of situations to improve MDA. PROGNOS aims especially to identify threatening targets (e.g., terrorist-ships). PROGNOS contains a

Probabilistic Ontology (PO) [Carvalho et al., 2017] based PROGNOS MTheory to support MDA, called PROGNOS PO.

Park et al. [2016] introduced an extended PROGNOS PO derived from the PSAW-MEBN reference model (Section 4), to develop more comprehensive PROGNOS PO. In the following, the extended PROGNOS PO is introduced.

### A) Extended PROGNOS PO developed from the PSAW-MEBN Reference Model

The original PROGNOS PO can be found in [Costa et al., 2012]. The PROGNOS PO containing five groups of MFrags. The first set of MFrags is to support of reasoning about a ship of interest. It includes nine MFrags *Aggressive Behavior*, *Terrorist Plan*, *Evasive Behavior*, *Erratic Behavior*, *Unusual Route*, *Bomb Port Plan*, *Ship Of Interest*, *Electronics Status*, and *Exchange Illicit Cargo Plan*. These MFrags are used to reason about properties of a ship (e.g., unusual behavior and an illegal plan). The second set of MFrags represents probabilistic knowledge about a person of interest. It includes four MFrags *Person Communications*, *Person Background Influences*, *Person Cluster Associations*, and *Person Relations*. These MFrags are used to identify a person who may communicate with a terrorist, has a suspicious background and history, and has a relationship with a terrorist. The third set of MFrags expresses information of relationships between two ships. It includes two MFrags *Radar* and *Meeting*. These MFrags are used to identify whether one ship is within radar range of another ship and whether two ships are meeting. The fourth set of MFrags is for information about the relationship between a person and an organization. It includes one MFrag *Terrorist Person* in which a person who belongs to an organization is identified. The last set of MFrags represents information about a relationship between a person and a ship. It includes two MFrags *Has Terrorist Crew* and *Ship Characteristics*. These MFrags are used to link a person and a ship, and to identify whether a ship has a terrorist crew member.

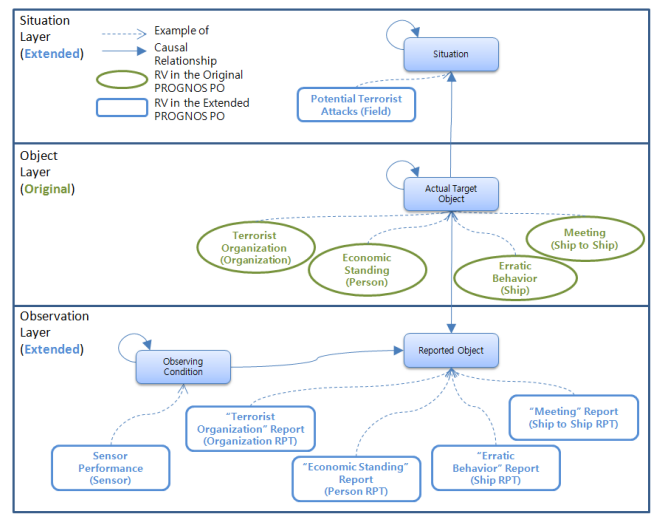

**Fig. 12 – Example RVs in the extended PROGNOS PO derived using the PSAW-MEBN reference model**

Park et al. [2016] introduced an extended PROGNOS PO using the PSAW-MEBN reference model. The original PROGNOS PO only dealt with the part of the object layer in Fig. 12. The extended PROGNOS PO contains the object layer as well as the situation layer and observation layer. In Fig. 12, the situation layer contains *Situation* RVs (e.g., the RV *Potential Terrorist Attacks* for a field). The observation layer contains *Observing Condition* RVs (e.g., the RV *Sensor Performance* for a sensor) and *Reported Object* RVs (e.g., the RV *Terrorist Organization Report* for an organization report).

Park et al. [2016] showed feature comparison between the original PROGNOS PO and the extended PROGNOS PO. The feature of the extended PROGNOS PO is more comprehensive than the feature of the original PROGNOS PO. The original PROGNOS PO contains 51 RVs, while the extended PROGNOS PO contains 115 RVs. This means that the extended PROGNOS PO can answer more various questions. For example, a reasoning about potential terrorist attacks in a field can be performed using the extended PROGNOS PO, but the original PROGNOS PO can't. Also, the extended PROGNOS PO contains observing conditions for sensors, so this may enable us to perform more accurate reasoning.

### B) Discussion

This research applied the PSAW-MEBN reference model to the original PROGNOS PO to identify whether we can improve the original model in terms of comprehensiveness. This research introduced that one can easily find lacks of a model supporting PSAW and develop more complete models via the PSAW-MEBN reference.

## 6. Conclusion

Previously, a PSAW-MTheory had to be designed from scratch. Designing PSAW-MTheories without any reference model is inefficient. To address this issue, this research presented a PSAW-MEBN reference model which can address questions of interest for PSAW. The PSAW-MEBN reference model was derived from a concept of PSAW in OODA. For this, we discussed the properties of PSAW and the properties of OODA. We presented the reference entities, the reference RVs, the reference causal relationship between RVs, and the reference MFrags for the PSAW-MEBN reference model. The PSAW-MEBN reference model could support the design of a PSAW-MTheory and improve the quality of the PSAW-MTheory.

In future work, the reference should be evaluated in other use cases of PSAW. Thereby, the PSAW-MEBN reference model may evolve into a more thorough reference model to address more complex situations. Also, the presented model should be extended to address other questions related to PSAW other than the questions addressed in this research.

### Acknowledgements

The research was partially supported by the Office of Naval Research (ONR), under Contract#: N00173-09-C-4008. We appreciate Dr. Paulo Costa and Mr. Shou Matsumoto for their helpful comments on this research.

## Appendix A. PSAW Questions

Roy [2001] proposed a broad spectrum of questions to be answered in situation analysis (e.g., situation element acquisition, data alignment and association, situation element perception refinement, situation element contextual analysis, situation element interpretation, situation classification/recognition, situation assessment, situation element projection, impact assessment, situation watch, and process refinement [Roy, 2001]). The following PSAW questions are adapted from the ones proposed by [Roy, 2001].

**Situation element acquisition**

(1) Does a (grouped) target exist?

(2) Is the (grouped) target identical to the previous detected (grouped) target?

(3) What are the environmental conditions?

**Data alignment and association**

(1) Is the target's information aligned with the common spatiotemporal space?

(2) Do the information items pertain to the same target or different targets?

**Situation element perception refinement**

(1) How many targets are moving?

(2) What size is the target?

(3) What type is the target?

(4) What group does the target belong to?

(5) What property of the target is not detected?

**Situation element contextual analysis**

(1) What is a relationship between the targets?

(2) What is the grouped target composed of?

(3) What causal relationships exist between the targets?

(4) How are activities related to each other?

**Situation element interpretation**

(1) What is the target doing?

(2) How often does an event take place?

(3) What are the aim, goal, objective, plan, and purpose of the target?

**Situation classification/recognition**

(1) What is the situation in which the targets are involved?

(2) What is the previous or next situation from the current situation?

(3) Has the situation existed previously?

**Situation assessment**

(1) How important is the situation?

(2) How dangerous is the situation?

(3) What is a remarkable event in the situation?

## Situation element projection

(1) Where are the (grouped) targets located in the future?

(2) What are the next activities performed by the (grouped) targets?

(3) When will be the activities from the (grouped) targets done?